\documentclass{article} %
\usepackage{iclr2022_conference}
\usepackage{times}

\usepackage{physics}

\usepackage{amsmath,amsfonts,bm}

\def\eqref#1{equation~\ref{#1}}

\def\1{\bm{1}}

\def\vone{{\bm{1}}}

\def\vf{{\bm{f}}}

\def\vk{{\bm{k}}}
\def\vK{{\bm{K}}}

\def\vo{{\bm{o}}}

\def\vq{{\bm{q}}}

\def\vs{{\bm{s}}}

\def\vv{{\bm{v}}}
\def\vw{{\bm{w}}}

\def\mA{{\bm{A}}}

\def\mF{{\bm{F}}}

\def\mK{{\bm{K}}}

\def\mO{{\bm{O}}}

\def\mQ{{\bm{Q}}}

\def\mV{{\bm{V}}}
\def\mW{{\bm{W}}}

\DeclareMathAlphabet{\mathsfit}{\encodingdefault}{\sfdefault}{m}{sl}
\SetMathAlphabet{\mathsfit}{bold}{\encodingdefault}{\sfdefault}{bx}{n}

\def\gF{{\mathcal{F}}}

\def\gO{{\mathcal{O}}}

\def\gW{{\mathcal{W}}}

\newcommand{\R}{\mathbb{R}}

\usepackage{todonotes}
\usepackage{url}
\usepackage[all]{foreign}
\usepackage{array}
\usepackage{graphicx}
\usepackage{natbib}
\usepackage{subcaption}
\usepackage{wrapfig}
\usepackage{makecell}
\usepackage{adjustbox}
\usepackage{epstopdf}
\usepackage{pgfplots}
\usepackage{multicol,multirow}
\usepackage{longtable}
\usepackage{enumitem}
\usepackage[ruled,vlined,linesnumbered,noend]{algorithm2e}
\usepackage{wrapfig}
\usepackage{float}
\usepackage[hidelinks]{hyperref}       %
\usepackage{booktabs}       %

\newcommand{\labelfont}{\it}

\tikzset{every plot/.append style={line width=0.7pt}}
\pgfplotsset{every axis/.append style={mark size=2}}
\pgfplotsset{every axis/.append style={font={\small}}}

\newcommand{\wrt}{w.r.t.~}
\renewcommand{\vs}{\textit{vs.}\ }

\renewcommand{\eqref}[1]{(\ref{#1})}

\def\myS~{\S{}}

\title{Token Pooling in Vision Transformers}

\author{Dmitrii Marin\thanks{Work done during internship at Apple.} \\ %
School of Computer Science, University of Waterloo \\
\texttt{dmitrii.marin@uwaterloo.ca} \vspace{-3mm}\\
\And
\And
Jen-Hao Rick Chang, Anurag Ranjan, Anish Prabhu, Mohammad Rastegari, Oncel Tuzel  \\
Apple \\
\texttt{\{jenhao\_chang,anuragr,anish\_prabhu,mrastegari,otuzel\}@apple.com} 
}

 \iclrfinalcopy %
\begin{document}

\maketitle

\begin{abstract}

Despite the recent success in many applications, the high computational requirements of vision transformers limit their use in resource-constrained settings.
While many existing methods improve the quadratic complexity of attention, in most vision transformers, self-attention is not the major computation bottleneck, \eg more than 80\% of the computation is spent on fully-connected layers. 
To improve the computational complexity of \textit{all} layers, we propose a novel token downsampling method, called \textit{Token Pooling}, efficiently exploiting redundancies in the images and intermediate token representations.
We show that, under mild assumptions, softmax-attention acts as a high-dimensional low-pass (smoothing) filter. 
Thus, its output contains redundancy that can be pruned to achieve a better trade-off between the computational cost and accuracy. 
Our new technique accurately approximates a set of tokens by minimizing the reconstruction error caused by downsampling. We solve this optimization problem via cost-efficient clustering. 
We rigorously analyze and compare to prior downsampling methods.
Our experiments show that Token Pooling significantly improves the cost-accuracy trade-off over the state-of-the-art downsampling.
Token Pooling is a simple and effective operator that can benefit many architectures.
Applied to DeiT, it achieves the same ImageNet top-1 accuracy using 42\% fewer computations.

\end{abstract}

\section{Introduction}

Vision transformers \citep{dosovitskiy2020image, touvron2020deit, liu2021swin, heo2021pit, zheng2021rethinking} have demonstrated state-of-the-art results in many vision applications, from image classification to segmentation. 
However, the high computational cost limits their use in resource-restricted, real-time, or low-powered applications.
While most prior work in Natural Language Processing (NLP) improve the time-complexity of attention \citep{tay2020efficient,ilharco2020high}, in vision transformers the main computation bottleneck is the fully-connected layers, as we show in \autoref{sec: complexity analysis}.  %
The computational complexity of these layers is determined by the number of tokens and their feature dimensionality. 
While reducing the dimensionality improves computational cost, it sacrifices model capacity and often significantly deteriorates the accuracy of the model.
On the other hand, since images often contain mostly smooth surfaces with sparsely located edges and corners, they contain similar (and thus redundant) features.
Moreover, we show that, under mild assumptions, softmax-attention is equivalent to low-pass filtering of tokens and thereby produces tokens with similar features, as empirically observed by \citet{goyal2020power} and \citet{Rao2021DynamicViTEV}.
This redundancy in representations suggests that we can reduce the number of tokens, \ie downsampling, without a significant impact to the accuracy, achieving a better cost-accuracy trade-off than reducing feature dimensionality alone.

Downsampling is widely used in Convolutional Neural Network (CNN) architectures to improve computational efficiency, among other purposes.
Given a grid of pixels or features, downsampling gradually reduces the grid dimensions via combining neighboring vertices on the grid.
The prevailing max/average pooling and sub-sampling are examples of (spatially uniform) \textit{grid-downsampling} that only uses locations on the grid to decide which vertices to combine. 
Such methods do not efficiently address non-uniformly distributed redundancy in images and features~\citep{recasens2018learning, Marin_2019_ICCV}.
Unlike CNNs that require grid preservation, transformers allow a wider range of nonuniform data-aware downsampling layers, where a better operator can be designed.

\begin{figure}[t]
	\centering
	\hspace{3.6mm}
	\begin{subfigure}[t]{0.32\linewidth}
		\centering
		\includegraphics[width=\linewidth] {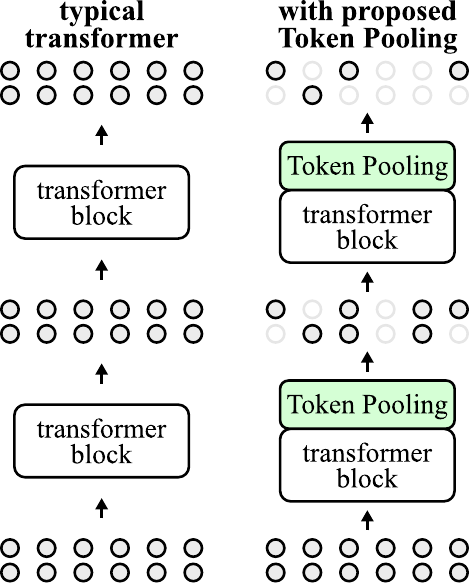}
		\caption{Proposed Token Pooling}
		\label{fig:arch}
	\end{subfigure}
	\hspace{5mm}
	\begin{subfigure}[t]{0.60\linewidth}
		\centering
		\includegraphics[width=0.85\linewidth] {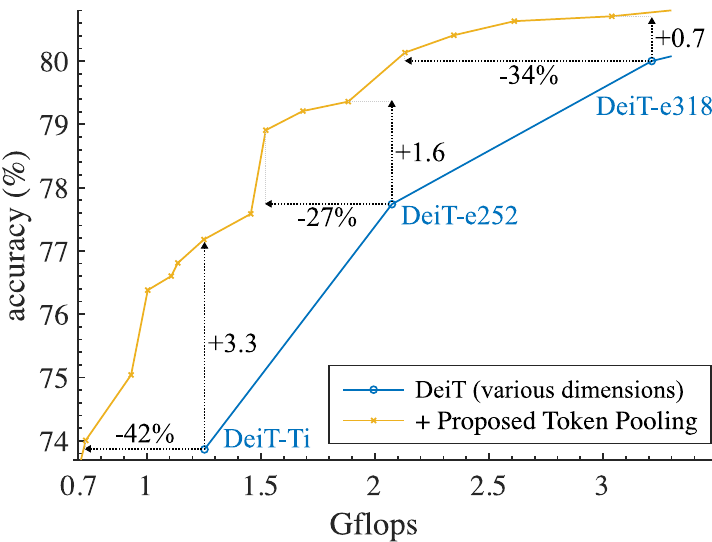}
		\caption{Accuracy \vs computation}
		\label{fig:teaser flop}
	\end{subfigure} 
	\\
	\vspace{2mm}
	\begin{subfigure}[t]{\linewidth}
		\centering
		\includegraphics[width=0.95\linewidth]{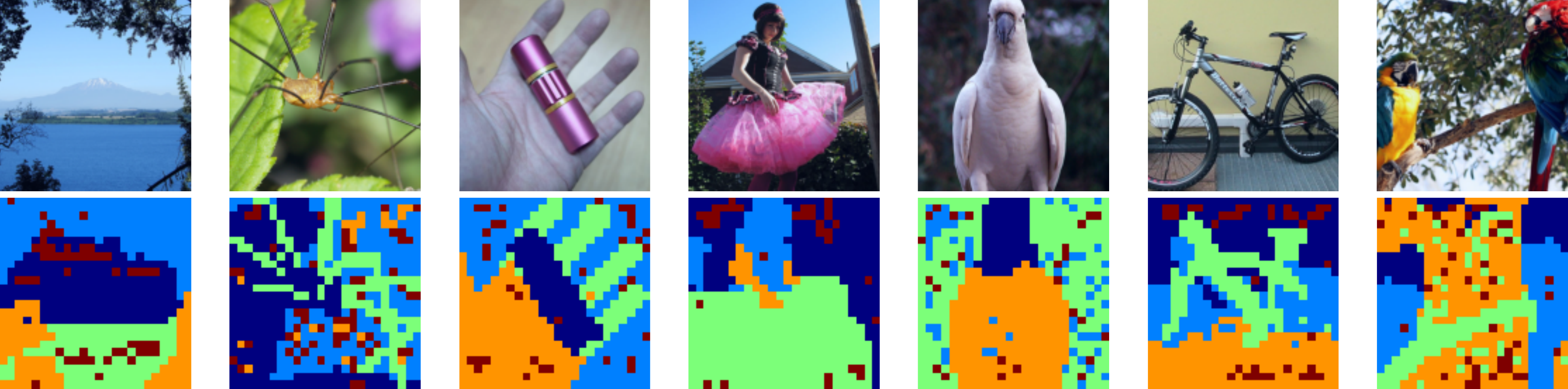}
		\caption{
			Token Pooling via cluster analysis of token representations.
		}
		\label{fig:teaser clusters}
	\end{subfigure} 
	\vspace{-2.6mm}
	\caption{(a) We propose Token Pooling, a novel token downsampling method, for visual transformers. (b) The proposed method achieves a state-of-the-art trade-off between accuracy and computation. (c) Token Pooling uses cluster analysis to aggregate information from individual tokens automatically. We show the input images and the token clusters at the 6-th layer of DeiT-S.}
	\label{fig:teaser}
\end{figure}

We propose \textit{Token Pooling}, a novel nonuniform data-aware downsampling operator for transformers efficiently exploiting redundancy in features. 
See the illustration and performance metric in Figures~\ref{fig:arch} \& \ref{fig:teaser flop}.
Motivated by nonuniform sampling and image compression~\citep{marvasti2012nonuniform,unat2009adaptive,belfor1994spatially,rabbani2002jpeg2000}, we formulate token downsampling as an optimization problem that minimizes the reconstruction error caused by downsampling. 
We show that clustering algorithms, K-Means and K-Medoids, efficiently solve this problem, see illustration in \autoref{fig:teaser clusters}.
To the best of our knowledge, we are the first to use this formulation and simple clustering analysis for token donwsampling in transformers.
We also compare with various prior downsampling techniques, including grid-downsampling \citep{pan2021scalable} and token pruning \citep{goyal2020power, Rao2021DynamicViTEV}.
Our results show that the proposed Token Pooling outperforms existing methods and provides the best trade-off between computational cost and classification accuracy.

\paragraph{Contributions.}  
The paper makes the following contributions: 
 \begin{itemize}[leftmargin=*, itemsep=0.11em, topsep=0.0em]
	\item We conduct an extensive study of prior downsampling techniques for visual transformers by comparing their computation-accuracy trade-offs.
	\item We analyze the computational cost of vision-transformer components and the limitations of the prior score-based downsampling methods. We also show that attention layers behave like low-pass filtering and thus produce redundant tokens. 
	\item Motivated by the redundancy in images and features, we propose a novel token downsampling technique, Token Pooling, for transformers via error minimization and achieve a significant improvement in the computation-accuracy trade-off. 
\end{itemize}

\section{Related work}
\label{sec:background}
In this section, we introduce vision transformers, and review existing methods that improve the efficiency of transformers including existing token-downsampling methods.

\subsection{Vision transformers}

Vision transformers \citep{dosovitskiy2020image, touvron2020deit, heo2021pit, liu2021swin, pan2021scalable} utilize the transformer architecture that is originally designed for NLP by \citet{vaswani2017attention} and further popularized by \citet{radford2018improving} and \citet{devlin2019bert}. 
In a high level, a vision transformer is a composition of $L$ transformer blocks that take a set of input tokens and return another set of output tokens.
In vision, input tokens are features representing individual non-overlapping image patches. %
To perform classification, a classification token is inserted to estimate the probabilities of individual classes.
To achieve the state-of-the-art, ViT \citep{dosovitskiy2020image} used pretraining on JFT-300M, a proprietary dataset much larger than standard ImageNet1k \citep{deng2009imagenet}.
Recently, DeiT \citep{touvron2020deit} achieved state-of-the-art results with advanced training on ImageNet1k only.

\begin{wrapfigure}{r}{15mm}
\vspace{-9mm}
\includegraphics[width=\linewidth]{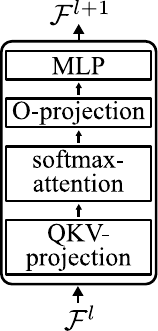}
\vspace{-11mm}
\end{wrapfigure}
\newcommand{\imF}{\mF}
\newcommand{\omF}{\tilde\mF}
Let the set of tokens at depth $l$ be $\gF^l=\{\vf^l_0, \dots, \vf^l_N\}$ where $\vf^l_i \in \R^M$ is the feature values of the $i$-th token. %
A typical transformer block $\phi$ at depth $l$ processes $\gF^l$ by a Multi-head Self-Attention (MSA) layer and a point-wise Multi-Layer Perceptron (MLP).
Let matrix $\imF \in \R^{N \times M} $ be a row-wise concatenation of tokens $\gF^l$.
Then\footnote{For compactness, we omit layer norm and skip-connections, see \citet{dosovitskiy2020image} for details.},
\begin{align}
    \phi(\imF) &\;=\; \operatorname{MLP}(\operatorname{MSA}(\imF)) \;\;\;\;\;\;\text{such that} \label{eq:MLP}\\
    \operatorname{MSA}(\imF) &\;=\; [\mO_1, \mO_2, \dots, \mO_H] \mW^O \label{eq:o proj}
\end{align}
where $H$ is the number of heads, matrix $\mW^O \in \R^{M \times M}$ is a learnable parameter of the block, $[,]$ is column-wise concatenation and $\mO_h \in \R^{N \times d}$ is the output of $h$-th attention head for $d=M/H$:
\begin{align}
    \mO_h &= \mA_h \mV_h \;\;\;\text{ such that }\;\;\; \mA_h = \operatorname{softmax}(\mQ_h \mK_h^\top {\big/} \sqrt{d}) \in \R^{N\times N}. \label{eq:msa}
\end{align}
Keys $\mK_h$, queries $\mQ_h$ and values $\mV_h$ are linear projections of the input tokens (QKV projections):
\begin{align}
    \mQ_h &= \imF \mW_h^Q, \;\;\;\;  \mK_h = \imF \mW_h^K, \;\;\;\; \mV_h = \imF \mW_h^V \label{eq:qkv}
\end{align}
where $\mW_h^Q \in \R^{M \times d}$, $\mW_h^K \in \R^{M \times d}$, $\mW_h^V \in \R^{M \times d}$ are learnable linear transformations. Note, the number of tokens is not affected by the transformer blocks, \ie $|\gF^{l+1}| = |\gF^l|$.

\subsection{Efficient transformers}\label{sec:
efficient transformers}

Similar to many machine learning models, the efficiency of transformers can be improved via 
meta-parameter search \citep{howard2017mobilenets,tan2019efficientnet}, 
automated neural architecture search \citep{elsken2019neural,tan2019mnasnet,wu2019fbnet}, manipulating the input size and resolution of feature maps \citep{paszke2016enet, howard2017mobilenets,zhao2018icnet}, pruning \citep{lecun1990optimal}, quantization \citep{jacob2018quantization}, and sparsification \citep{gale2019state}, \etc.
For example,  \citet{dosovitskiy2020image} and \citet{touvron2020deit} obtain a family of ViT and DeiT models, respectively, by varying the input resolution, the number of heads $H$, and the feature dimensionality $M$. 
Each of the models operates with a different computational requirement and accuracy.
In the following, we review techniques developed for transformers.

\subsubsection{Efficient self-attention}

The softmax-attention layer \eqref{eq:msa} has a quadratic time complexity \wrt the number of tokens, \ie $\mathcal{O}(N^2)$. 
In many NLP applications where every token represents a word or a character, $N$ can be large, making attention a computation bottleneck \citep{dai2019transformer,rae2019compressive}.
While many works improve the time complexity of attention layers, as we will see in \autoref{sec:bottleneck}, they are not the bottleneck in most current vision transformers.

The time complexity of an attention layer can be reduced by restricting the attention field-of-view and thus imposing sparsity on $\mA_h$ in \eqref{eq:msa}. 
This can be achieved using the spatial relationship between tokens in the image/text domain \citep{parmar2018image, ramachandran2019stand, qiu2020blockwise, Beltagy2020Longformer, child2019generating, zaheer2020big} or based on token values using locality-sensitive hashing, sorting, compression, \etc \citep{kitaev2020reformer, roy2021efficient, vyas2020fast, tay2020sparse, liu2018generating, wang2020linformer, tay2021synthesizer}.
Prior works have also proposed attention mechanisms with lower time complexity, \eg $\mathcal{O}(N)$ or $\mathcal{O}(N \log N)$~\citep{katharopoulos2020transformers, peng2021random, choromanski2021rethinking, tay2021synthesizer}.

Note that the goal of these methods is to reduce the time complexity of the attention layer---the number of tokens remains the same across the transformer blocks.
In contrast, our method reduces the number of tokens \textit{after} attention has been computed.
Thereby, we can utilize these methods to further improve the overall efficiency of transformers.

Recently, \citet{wu2020visual} proposed a new attention-based layer that learns a small number of query vectors to extract information from the input feature map.
Similarly, \citet{wu2021centroid} replace self-attention with a new recurrent layer that outputs a smaller number of tokens.
In comparison, our method directly minimizes the token reconstruction error due to token downsampling.
Also, our layer has no learnable parameters and can be easily incorporated into existing vision transformers.

\subsubsection{Downsampling methods for transformers}
\label{sec:token downsampling methods}

\paragraph{Grid-downsampling.}

The input tokens of the first vision transformer block are computed from image patches. 
Therefore, even though transformers treat tokens as a set, we can still associate a grid to the tokens using their initial locations on the image.
The regular grid structure allows typical downsampling methods, such as max/mean pooling, uniform sub-sampling. 
For example, \citet{liu2021swin, heo2021pit} use convolutions with stride to downsample the feature maps formed by the tokens. Also, these works increase the feature dimensionality with depth, similarly to CNNs.

\paragraph{Score-based token downsampling.}
\label{sec:score methods}

In the area of NLP, \citet{goyal2020power} introduce the idea of dropping tokens.
Their approach, called PoWER-BERT, is based on a measure of \textit{significance score}, which is defined as the total attention given to a token from all other tokens.
Specifically, the significance scores of all tokens in the $l$-th transformer block, $\bm{s}^l \in \R^N$, is computed by
\begin{equation}\label{eq:significance}
     \bm{s}^l = \sum_{h=1}^H {\mA_h^l}^\top \vone,
\end{equation}
where $\mA_h^l$ is the attention weights of head $h$ defined in \eqref{eq:msa}.
They only pass $K_l$ tokens with the highest scores in $\bm{s}^l$ to the next transformer block.
The pruning is performed on all blocks.

PoWER-BERT is trained using a three-stage process. 
First, given a base architecture, they pretrain a model without pruning.
In the second stage, a soft-selection layer is inserted after each transformer block, and the model is finetuned for a small number of epochs. 
Once learned, the number of tokens to keep, $K_l$, for each layer is computed from the soft-selection layers.
Last, the model is finetuned again with the tokens pruned using the $K_l$ from the second stage.
See \citet{goyal2020power} for details.

Recently, \citet{Rao2021DynamicViTEV} proposed Dynamic-ViT that also uses scores to prune tokens. 
Unlike PoWER-BERT, which computes significance scores from attention weights, \citeauthor{Rao2021DynamicViTEV} use a dedicated sub-network with learned parameters. 
The method requires knowledge distillation, Gumbel-Softmax, and straight-through estimators on top of the DeiT training.

We will analyze the limitations of score-based methods in \autoref{sec:limitations of scores}.

\section{Analysis}

This section addresses three questions. 
First, it identifies the computational bottleneck of vision transformers. 
Second, it discusses the limitations of score-based downsampling. %
Third, it analyzes how the softmax-attention affects the redundancy in tokens.

\subsection{Computation analysis of vision transformers}
\label{sec: complexity analysis}
\label{sec:bottleneck}

\begin{table}[t]
    \centering
    \begin{adjustbox}{max width=0.95\linewidth}
    \begin{tabular}{l|c|cccc}
        \makecell{Layer} & \makecell{Complexity} & \multicolumn{4}{c}{Computation ($10^9$ Flops)} \\
         & & \makecell{ViT-B/384 \\ ($N = 577$)}  & \makecell{ViT-B \\ ($N = 197$)} & \makecell{DeiT-S \\ ($N = 197$)} & \makecell{DeiT-Ti \\ ($N = 197$)} \\
        \hline
        softmax-attention &
        $\gO(LN^2M)$ & 6.18 & 0.72 & 0.36 & 0.18 \\
        QKV projections &
         $\gO(LNM^2)$  & 12.25 & 4.18 & 1.05 & 0.26\\
        O projection &
        $\gO(LNM^2)$ & 4.08 & 1.39 & 0.35 & 0.09 \\
        Multi-linear Perceptron &
        $\gO(LNM^2)$ & 32.67 & 11.15 & 2.79 & 0.70 \\
        \hline
        Total & $\gO(LNM(M+N))$ & 55.5 & 17.6 & 4.6 & 1.3 \\
    \end{tabular}
    \end{adjustbox}
    \\
    \begin{subfigure}[t]{\linewidth}
        \centering
        \vspace{3.6mm}
        \includegraphics[width=0.95\linewidth]{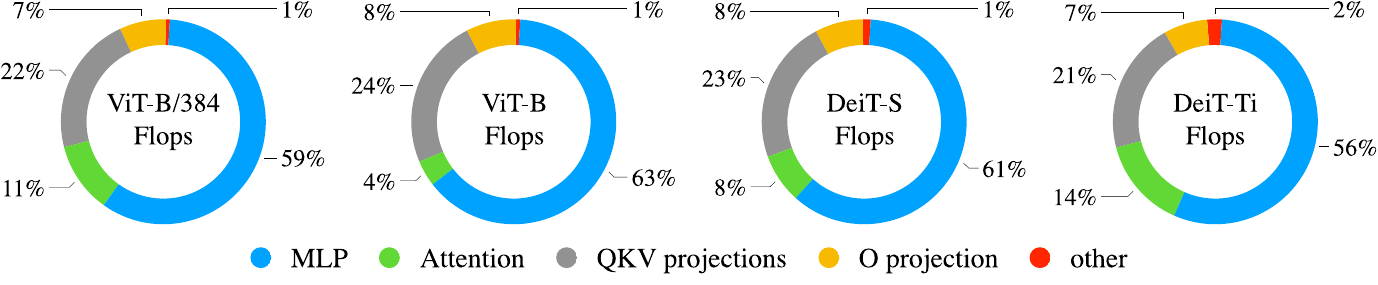}
    \end{subfigure}
    \vspace{-2mm}
    \caption{Time complexity and computation breakdown of ViT \citep{dosovitskiy2020image} and DeiT \citep{touvron2020deit}. $L$ is the number of transformer blocks, $N$ is the number of input tokens (patches), and $M$ is the feature dimensionality.  All models take input images of size $224\times224$ except ViT-B/384, which uses $384\times384$. The softmax-attention layers constitute a fraction (15\% or less) of the total compute, whereas fully-connected layers (MLP and projections) spend over 80\%.
    }
    \label{tab:complexity and flops}
    \label{fig:flops distribution}
\end{table}

We analyze the time complexity and computational costs (measured in flops) of commonly used vision transformers, namely ViT and DeiT.
We breakdown the computation into four categories: softmax-attention \eqref{eq:msa}, QKV projections \eqref{eq:qkv}, O projection \eqref{eq:o proj} and MLP \eqref{eq:MLP}. 
As shown in \autoref{tab:complexity and flops}, in all these vision transformers, the main computational bottleneck is the fully-connected layers that spend over 80\% of the total computation.
In comparison, softmax-attention only takes less than 15\%.
Note that we explicitly break down the multi-head attention into the softmax-attention, QKV and O projections, as they have different time complexity, see \autoref{tab:complexity and flops}.
This decomposition reveals that the QKV and O projections spend most of the computations of the multi-head self-attention.

\subsection{Limitations of score-based token downsampling}
\label{sec:limitations of scores}

\begin{figure}[t]
    \centering
    \vspace{3.6mm}
    \begin{subfigure}[t]{0.45\linewidth}
        \centering
        \includegraphics[width=\linewidth]{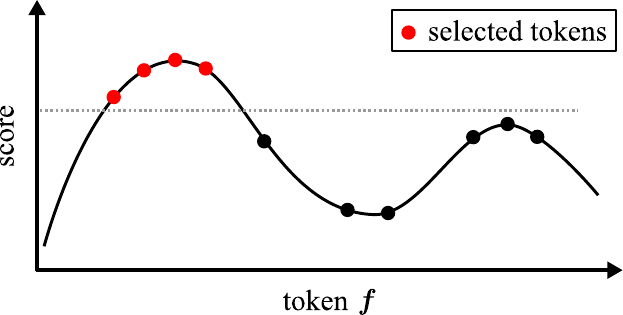}
        \caption{Score-based downsampling}
        \label{fig:score vs clustering:score}
    \end{subfigure}
    \hspace{8mm}
    \begin{subfigure}[t]{0.45\linewidth}
        \centering
        \includegraphics[width=\linewidth]{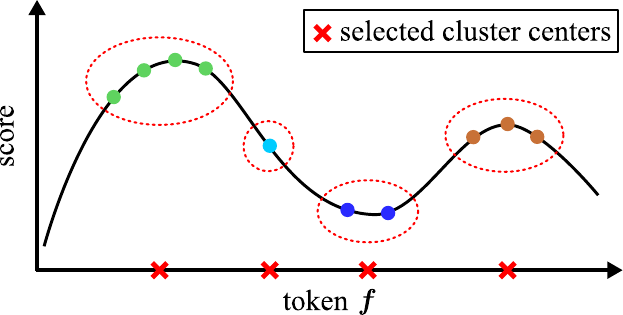}
        \caption{Proposed downsampling}
        \label{fig:score vs clustering:clustering}
    \end{subfigure}
    \vspace{-1.6mm}
    \caption{Score-based downsampling methods \citep{goyal2020power,Rao2021DynamicViTEV} \vs our method.  
    In the figure, the x-axis represents the token values (in one dimension), and the y-axis represents their scores. Suppose four tokens are to be selected. 
    (a) Score-based methods select tokens with higher scores. Since the scoring function is continuous, all tokens in the left lobe will be selected, resulting in redundancy and information loss in the right lobe. 
    (b) The proposed method first forms four clusters to approximate the set of tokens, then selects the cluster centers. Thus, the output tokens are a more  accurate representation of the original token set than the score-based methods.}
    \label{fig:score vs clustering}
\end{figure}

Existing score-based token downsampling methods like PoWER-BERT and Dynamic-ViT utilize scoring functions to determine the tokens to keep (or prune).
They keep tokens with the top-K scores and discard the rest. 
Since these scoring functions are continuous with limited Lipschitz constants, tokens that are close in the feature space will be assigned similar scores.
Therefore, these similar tokens will likely either be all kept or discarded, as illustrated in \autoref{fig:score vs clustering:score}.
As our experiments show, this redundancy (in the kept tokens) and severe information loss (in the pruned tokens) deteriorate the computation-accuracy trade-off of the score-based downsampling methods.

\subsection{Attention as a low-pass filter}
\label{sec:low-pass} 

Given a query vector $\vq$, a set of key vectors $\mathcal{K} = \{ \vk_1, \dots, \vk_N\}$, the corresponding value vectors $\mathcal{V} = \{\vv_1, \dots, \vv_N \}$ and a scalar $\alpha > 0$, softmax-attention computes the output via
\begin{align}
    \vo(\vq) = \frac{1}{z(\vq)} \sum_{i=1}^N \exp(\alpha \, \vq \cdot \vk_i) \, \vv_i && \mbox{ where } && z(\vq) = \sum_{i=1}^N \exp(\alpha \, \vq \cdot \vk_i).
    \label{eq:attention}
\end{align}
Note that we write $\vo(\vq)$ to indicate that the output vector $\vo$ is a function of the query $\vq$. 
If the query vector and all key vectors are normalized to have a fixed $\ell^2$ norm, we can rewrite \eqref{eq:attention} as 
\begin{align}
    \vo(\vq) & 
    = \frac{1}{z'(\vq)} \! \sum_{i=1}^N \exp\!\left(-\frac{\alpha}{2} \| \vq - \vk_i \|^2  \right) \vv_i \nonumber 
    = \frac{1}{z'(\vq)} \! \int \! \exp\!\left(-\frac{\alpha}{2} \| \vq - \vk \|^2  \right) \! \left( \sum_{i=1}^N \! \delta(\vk - \vk_i) \vv_i \right) \dd{\vk} \nonumber \\ 
    & = \frac{1}{z'(\vq)} \ G\!\left(\vq; \frac{1}{\alpha}\right) * S(\vq; \mathcal{K}, \mathcal{V}) %
    \label{eq:attention as filter} 
\end{align}
where $*$ represents high-dimensional convolution, $z'(\vq) = \sum_i \exp(-\frac{\alpha}{2} \|\vq - \vk_i\|^2) = G\!\left(\vq; \frac{1}{\alpha}\right) * S(\vq; \mathcal{K}, 1)$ is the normalization scalar function, $G\!\left(\vq; \sigma^2\right) = \exp({ -\| \vq \|^2}\big/{2 \sigma^2})$ is an isometric Gaussian kernel, and $S(\vq; \mathcal{K}, \mathcal{V}) = \sum_{i=1}^N \delta(\vq - \vk_i) \, \vv_i$ is a high-dimensional sparse signal, which is composed of $N$ delta functions located at $\vk_i$ with value $\vv_i$.
According to \eqref{eq:attention as filter}, given query vectors $\vq_1, \dots, \vq_N$, there are two conceptual steps to compute softmax-attention: 
\begin{enumerate}[leftmargin=8mm, topsep=0pt, itemsep=0ex, partopsep=1ex, parsep=1ex]
    \item filter $S(\vq; \mathcal{K}, \mathcal{V})$ with a Gaussian kernel to get $\vo(\vq)$, and
    \item sample $\vo(\vq)$ at coordinates $\vq_1, \dots, \vq_N$ to get the output vectors $\vo_1, \dots, \vo_N$.
\end{enumerate}
Since Gaussian filtering is low-pass, $\vo(\vq)$ is a smooth signal.
Therefore, the output tokens of the attention layer, \ie discrete samples of $\vo(\vq)$, contain similar feature values. 
So, based on Nyquist-Shannon sampling theorem~\citep{Shannon1949}, there exists redundant information in the output, which is used by our methods to reduce computation without losing much of the important information.

Note that our analysis is based on the normalized query and key vectors, which can be achieved by inserting a normalizing layer before the softmax-attention layer without significantly affecting the performance of a transformer, as shown by \citet{kitaev2020reformer}. %
It has also been empirically observed by \citet{goyal2020power} and \citet{Rao2021DynamicViTEV} that even without the normalization, transformers produce tokens with similar values.
To demonstrate this, in all our experiments, %
we use standard multi-head attention \textit{without} normalizing keys and queries.
We conduct an ablation study with normalized keys and queries in \autoref{app:normalized key query}.

\section{Token pooling}

Pruning tokens inevitably loses information. In this section, we formulate a new token downsampling principle enabling strategical tokens selection that preserves the most information. 
Based on this principle, we formulate and discuss several Token Pooling algorithms.

Given a set of output tokens $\gF = \{ \vf_1, \dots, \vf_N \}$ of a transformer block, our goal is to find a smaller set of tokens $\hat{\gF} = \{ \hat{\vf}_1, \dots, \hat{\vf}_K \}$ that after upsampling minimizes the reconstruction error of $\gF$.
Specifically, the reconstruction of $u(\vf_i; \hat{\gF})$ of token $\vf_i$ is computed via interpolating the tokens in~$\hat{\gF}$.
The reconstruction error is then defined as
\begin{equation}
    \ell(\gF, \hat{\gF}) = \sum_{\vf_i \in \gF} \| \vf_i - u(\vf_i; \hat{\gF}) \|^2.
    \label{eq:ori loss}
\end{equation}
To simplify our formulation and reduce computation, we use \textit{nearest-neighbor} interpolation as $u$.
As a consequence, the reconstruction error \eqref{eq:ori loss} becomes %
\begin{equation}
    \ell(\gF, \hat{\gF}) = \sum_{\vf_i \in F} \  \min_{\hat{\vf}_j \in \hat{F}} \ \| \vf_i - \hat{\vf}_j \|^2,
    \label{eq:reconstruction loss}
\end{equation}
which is the asymmetric Chamfer divergence between $\gF$ and $\hat{\gF}$ \citep{barrow1977parametric, mechrez2018maintaining}.
The loss \eqref{eq:reconstruction loss} can be minimized by the K-Means algorithm, \ie clustering the tokens in $\gF$ into $K$ clusters.

The proposed Token Pooling layer is defined in \autoref{al:downsampling}.
It downsamples input tokens via clustering the tokens and returns the cluster centers (the average of the tokens in a cluster).
As we have shown above, this operation directly minimizes the reconstruction error \eqref{eq:reconstruction loss} caused by the downsampling.
Intuitively, clustering the tokens provides a more accurate and diverse representation of the original set of tokens, compared to the top-K selection used by score-based downsampling methods, as shown in \autoref{fig:score vs clustering:clustering}.
Note that Token Pooling is robust to the initialization of cluster centers, as shown in \autoref{app:initialization}. 
Below, we provide details of the clustering algorithms.

\paragraph{K-Means.}

We use the K-Means algorithm to minimize \eqref{eq:reconstruction loss} via the following iterations:
\begin{align}
    a(i) & \;\;\; \leftarrow \;\;\; \arg\min_j \| \vf_i - \hat\vf_j\| && \forall i \in \{1,2,\dots, N\}  \label{eq:kmeans assignment} \\
    \hat\vf_j & \;\;\; \leftarrow \;\;\; \sum_{i=1}^N [a(i) = j] \vf_i \Bigg/ \sum_{i=1}^N  [a(i) = j]  && \forall j \in \{1,2,\dots, K\} \label{eq:kmeans centroid update}
\end{align} 
where $[\,]$ is the Iverson bracket and $a$ is the cluster assignment function. 
The overall algorithm complexity is $\gO(TKNM)$ where $T$ is the number of iterations. 
The vast majority of the computation is spent on the repetitive evaluation of the distances between tokens and centroids in step \eqref{eq:kmeans assignment}.

\paragraph{K-Medoids.}

We can use the more efficient K-Medoids algorithm by replacing step \eqref{eq:kmeans centroid update} with:
\begin{align}
    n(j) \; \leftarrow \; {\arg\min_{i:a(i) = j} \sum_{i':a(i')=j} \|\vf_i - \vf_{i'} \|^2} %
    && \text{and} &&
    \hat\vf_j  \; \leftarrow \; \vf_{n(j)}. \label{eq:kmedoids centroid update} %
\end{align}
These steps minimize objective \eqref{eq:reconstruction loss} under the \textit{medoid constraint}: $\hat \gF \subseteq \gF$.

The advantage of the K-Medoids algorithm is its time complexity $\gO(TKN + N^2M)$, which is substantially lower in practice as we only compute the distances between tokens once. 
In our experience, it requires less than 5 iterations to converge.
Apart from the distance matrix computation, the cost of the K-Medoids algorithm is negligible when compared with the cost of the other layers.

\begin{algorithm}[t]
\SetKwInOut{Input}{input}
\SetKwInOut{Output}{output}
\Input {tokens $\gF=\{\vf_1,\dots,\vf_N\}$; target set size $K$; (optional) weights $\gW=\{w_1,\dots, w_N\}$}
\Output {downsampled set $\hat \gF=\{\hat \vf_1,\dots,\hat \vf_K\}$}
\lIf{$k \ge N$}{\Return{F}}
initialize cluster centers $\hat \gF$ to be the $K$ tokens from $\gF$ with the highest weights \;
\While{not converged \emph{and} max number of iterations is not reached}{
    \lFor{ $i\in\{1,...,N\}$ }{ update cluster assignment $z(i) \leftarrow \arg\min_{j=1}^K \| \vf_i -\hat\vf_j\|$ }
    \lFor{ $j\in\{1,...,K\}$ }{ update cluster center $\hat{\vf}_j$ according to the chosen clustering algorithm, that is either \eqref{eq:kmeans centroid update}, \eqref{eq:kmedoids centroid update}, \eqref{eq:w kmeans centroid update} or \eqref{eq:w kmedoids centroid update}
    }
}
\Return{weighted means of tokens in each cluster}
\caption{Token Pooling}
\label{al:downsampling}
\end{algorithm}

\paragraph{Weighted clustering.}

Reconstruction error \eqref{eq:reconstruction loss} treats every token equally; however, each token contributes differently to the final output of an attention layer. %
Thus, we also consider a weighted reconstruction error: $
    \ell(\gF, \hat{\gF};
    \vw) = \sum_{\vf_i \in F} \min_{\hat{\vf}_j \in \hat{F}} w_i \| \vf_i - \hat{\vf}_j \|^2$
where $\vw = [w_1, \dots, w_N]$ are the positive weights corresponding to the individual tokens in $\gF$. \autoref{app:weighted clustering} details the clustering algorithms for the weighted case.
A good choice of the weights is the significance scores \eqref{eq:significance}, \ie $w_i = s_i$. %
The significance score identifies the tokens that influence the current transformer block most and thus should be approximated more precisely.

\section{Experiments} \label{sec:results}

Our implementation is based on DeiT \citep{touvron2020deit} where we insert downsampling layers after each transformer block.
To evaluate the pure effect of downsampling, we keep all meta-parameters of DeiT, including the feature dimensionality, network depth, learning rate schedule, \etc.
We also do not use knowledge distillation during training. 
We use ImageNet1k classification benchmark \citep{russakovsky2015imagenet}. 
Our cost and performance metrics are flops and top-1 accuracy.

\paragraph{Methods.}
We evaluate the following token downsampling methods:   
\begin{enumerate}[leftmargin=*]
    \item \textit{Convolution} downsampling. We implement uniform grid-downsampling via convolution with stride 2, \ie, the tokens corresponding to the adjacent image patches are concatenated and mapped to $\R^M$. This design is used in \citet{liu2021swin, heo2021pit}. See details in \autoref{sec:convolution}.
    \item \textit{PoWER-BERT.} We implement PoWER-BERT on DeiT, following \citep{goyal2020power}.
    \item \textit{Random selection}, a simple baseline randomly selecting $K$ tokens without replacement.
    \item \textit{Importance selection} chooses $K$ tokens by drawing from a categorical distribution without replacement with probabilities proportional to the significance score \eqref{eq:significance} of each token.
    \item \textit{Token Pooling} use K-Means or K-Medoids algorithms, or their weighted versions, WK-Means or WK-Medoids, respectively. The weights are the significance scores \eqref{eq:significance}. 
\end{enumerate}

\paragraph{Selection of $\vK=[K_1, \dots, K_L]$.}
To fairly compare our Token Pooling with PoWER-BERT and other baselines, all methods (except convolution downsampling) use the same number of  retained tokens for downsampling layers. \autoref{app:2x} details the selection of $\vK$.

\paragraph{Training protocol.}
(1)~A base DeiT model (\eg DeiT-S) is trained using the original training \citep{touvron2020deit}. 
(2)~We then finetune the base DeiT model using the second stage of PoWER-BERT's training and acquire $\vK$.
(3)~We further finetune the downsampling methods using $\vK$. 
(4)~We also finetune the base DeiT model. 
Our protocol ensures a fair comparison such that all of the models are trained with the same number of iterations, the same learning rate schedule, \etc.

\begin{figure}[t]
    \centering
    \begin{subfigure}[t]{0.48\linewidth}
        \centering 
        \includegraphics[width=\linewidth]{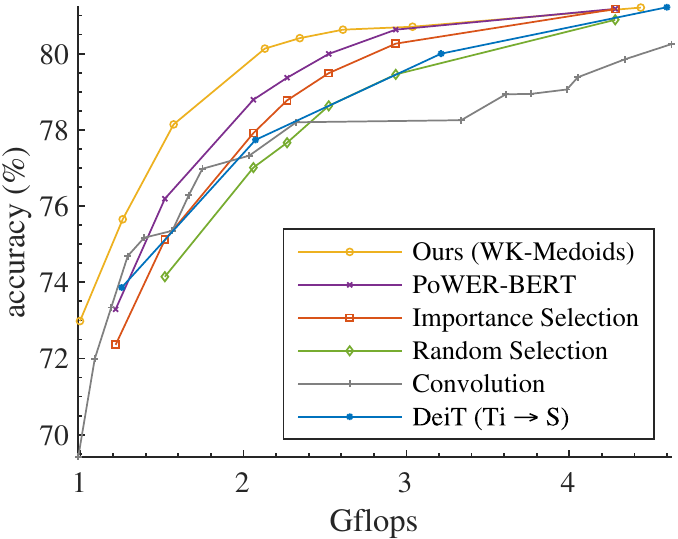}
        \caption{Models based on DeiT-S}
        \label{fig:deit-s}
    \end{subfigure}
    \hfill
    \begin{subfigure}[t]{0.48\linewidth}
        \centering 
        \includegraphics[width=\linewidth]{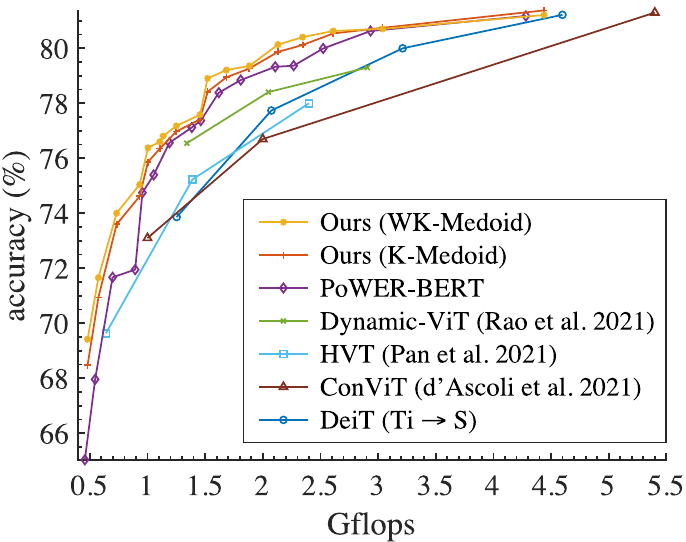}
        \caption{Comparison with current SOTA}
        \label{fig:sota}
    \end{subfigure}
    \caption{\textbf{Main results.} (a) shows the accuracy when we apply different downsampling methods to DeiT-S.  More is in \autoref{app:clustering ablations}. (b) shows a comparison between the proposed method with the state-of-the-art downsampling methods. The results of our method and PoWER-BERT are acquired by varying $\vK$ and the base architecture among DeiT-Ti, DeiT-e252, DeiT-e318, and DeiT-S.}
    \label{fig:main results}
\end{figure}

\subsection{Main results}

First, we apply different downsampling methods on DeiT-S. 
As shown in \autoref{fig:deit-s}, random selection achieves a similar trade-off as lowering feature dimensionality $M$. 
While convolution with stride is better than adjusting $M$ at the low-compute regime, it fails in high-compute regimes.
Importance selection improves upon random selection but is still outperformed by PoWER-BERT.
Our Token Pooling (with weighted K-Medoids) achieves the best trade-off in all regimes.

\begin{figure}[t]
    \centering
    \begin{minipage}{0.48\textwidth}
        \includegraphics[width=\linewidth]{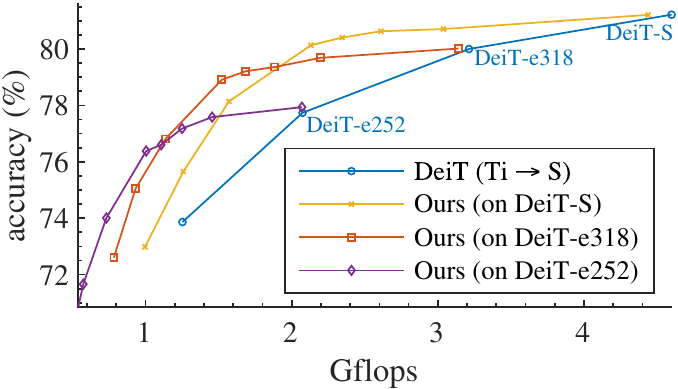}
    \end{minipage}%
    \hfill
    \begin{minipage}{0.48\textwidth}
        \caption{
        The figure shows the results when we apply Token Pooling to various DeiT architectures. Token Pooling consistently improves the computation-accuracy trade-off for all evaluated architectures. By utilizing both Token Pooling and architecture search, we can further improve the accuracy at a given flops budget. For example, at 1 Gflop, we should use Token Pooling on DeiT-e252 instead of DeiT-S.
        }
        \label{fig:width and clustering}
    \end{minipage}
\end{figure}

Next, we apply Token Pooling to DeiT models with different $M$ (DeiT-Ti, DeiT-e252, DeiT-e318, and DeiT-S).  
\autoref{fig:width and clustering} shows trade-off curves for each DeiT model. 
Token Pooling enables each of the models to achieve a better computation-accuracy trade-off than simply varying the feature dimensionality $M$.
For each computational budget, we find the combination of $M$ and $\vK$ that gives the highest accuracy.
The best balance is achieved by applying Token Pooling and selecting the best $M$.
We do the same for PoWER-BERT.
\autoref{fig:sota} shows the results of the proposed Token Pooling and state-of-the-art methods.
We cite the results of \citet{Rao2021DynamicViTEV}, \citet{d2021convit}, and \citet{pan2021scalable}.
Our Token Pooling achieves the best accuracy across all evaluated compute regimes.

Finally, we compare the best computation-accuracy trade-off achieved by Token Pooling (with WK-Medoids and varying $M$) with standard DeiT models in \autoref{fig:teaser flop}.
As can be seen, utilizing Token Pooling, we significantly improve the computation costs of DeiT-Ti by 42\% and improve the top-1 accuracy by 3.3 points at the same flops. 
Similar benefits can be seen on DeiT-e252 and DeiT-e318.

\subsection{Ablation studies}

\begin{figure}[t]
    \centering
    \begin{subfigure}[t]{0.48\linewidth}
        \centering
        \includegraphics[width=\linewidth]{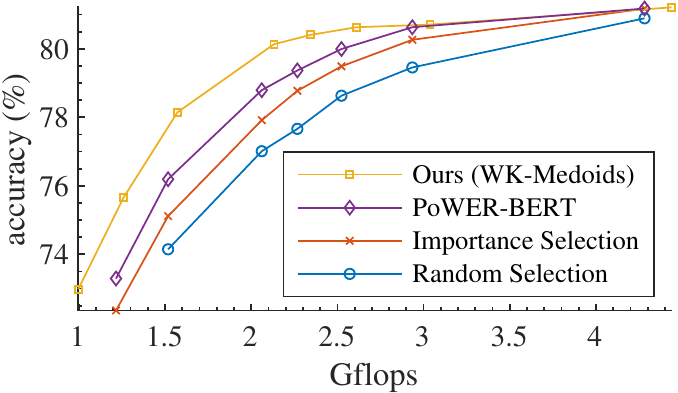}
        \caption{Ours \vs methods using significance score}
        \label{fig:score methods}
    \end{subfigure}
    \hfill
    \begin{subfigure}[t]{0.48\linewidth}
        \centering
        \includegraphics[width=\linewidth]{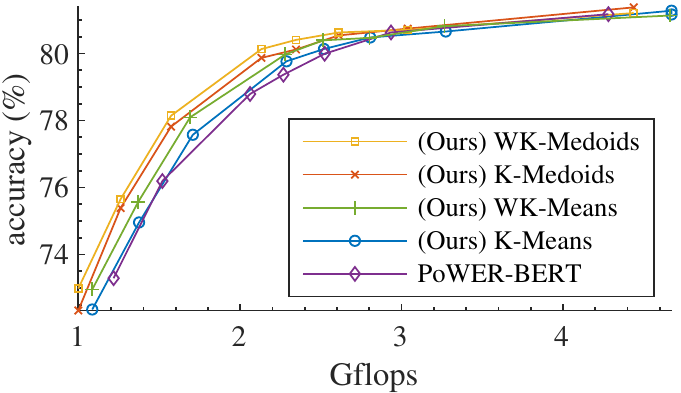}
        \caption{Variants of Token Pooling}
        \label{fig:token poolings}
    \end{subfigure}
    \caption{Ablation studies of (a) downsampling methods using significance score, and (b) proposed Token Pooling using different clustering algorithms. The base model is DeiT-S for all methods.}
    \label{fig:ablations}
\end{figure}

\autoref{fig:score methods} compares methods utilizing significance scores.
As can be seen, using importance selection improves upon the simple random selection.
By minimizing the reconstruction error \eqref{eq:reconstruction loss}, our method achieves better cost-accuracy trade-off. 
\autoref{fig:token poolings} evaluates Token Pooling with different clustering algorithms. Weighted versions outperform regular versions of K-Means and K-Medoids. 
Due to the higher time complexity, K-Means is outperformed by K-Medoids (the curves are shifted toward the right).
See the metrics and flops used by clustering  in table format in \autoref{app:clustering ablations}. 
Still, all Token Pooling variants outperform the baseline, demonstrating the effectiveness of our method. 

More ablation studies are in the appendices, including the effect of cluster-center initialization (\autoref{app:initialization}), results of convolution downsampling (\autoref{sec:convolution}), results of Token Pooling before finetuning (\autoref{app:no tuning}), results with normalized keys and queries of softmax attention (\autoref{app:normalized key query}), and detailed information ($\vK$, flops, accuracy, clustering cost) of our models (\autoref{app:clustering ablations}).

\section{Conclusions}

This paper provides two insights of vision transformers: 
first, their computational bottleneck is the fully-connected layers, and second, attention layers generate redundant representations due to the connection to  Gaussian filtering.
Token Pooling, our novel nonuniform data-aware downsampling operator, utilizes these insights and significantly improves the computation-accuracy balance of DeiT, compared to existing downsampling techniques. 
We believe that Token Pooling is a timely development and can be used with other techniques (\eg architecture search, quantization) to spur the design of efficient vision transformers.

\bibliography{main}
\bibliographystyle{iclr2022_conference}

\clearpage

\appendix

\section{Weighted clustering algorithms}
\label{app:weighted clustering}

\paragraph{Weighted K-Means}

minimizes the following objective \wrt $\hat\gF=\{\hat{\vf}_1, \dots, \hat{\vf}_K\} \subset \R^M$:
\begin{align}
    \ell(\gF, \hat{\gF}) = \sum_{\vf_i \in F} \  \min_{\hat{\vf}_j \in \hat{F}} w_i\| \vf_i - \hat{\vf}_j \|^2%
    \label{eq:weighted k means objective}
\end{align}
The extension of the K-Means algorithm to the weighted case iterates the following steps:
\begin{align}
    a(i) & \;\;\; \leftarrow \;\;\; \arg\min_j \| \vf_i - \hat\vf_j\| && \forall i \in \{1,2,\dots, N\}, \\
    \hat\vf_j & \;\;\; \leftarrow \;\;\;  \frac{\sum_{i=1}^N [a(i) = j] w_i\vf_i }{\sum_{i=1}^N  [a(i) = j]w_i}  && \forall j \in \{1,2,\dots, K\}. \label{eq:w kmeans centroid update}, 
\end{align}

\paragraph{Weighted K-Medoids} optimizes objective \eqref{eq:weighted k means objective} under the medoid constraint $\hat\gF \subset \gF$:
\begin{align}
    a(i) & \;\;\; \leftarrow \;\;\; \arg\min_j \| \vf_i - \hat\vf_j\| && \forall i \in \{1,2,\dots, N\}, \\
    n(j) & \;\;\; \leftarrow \;\;\; \operatornamewithlimits{\arg\min}_{i:\,z(i) = j} \sum_{i':\,a(i')=j} \|\vf_i - \vf_{i'} \|^2 && \forall j \in \{1,2,\dots, K\}, \\
    \hat\vf_j & \;\;\; \leftarrow \;\;\; \vf_{n(j)}. \label{eq:w kmedoids centroid update}
\end{align}

\section{Clustering initialization ablations}
\label{app:initialization}

\begin{figure}[H]
    \centering
    \begin{tikzpicture}

\definecolor{color0}{rgb}{0.12156862745098,0.466666666666667,0.705882352941177}
\definecolor{color1}{rgb}{1,0.498039215686275,0.0549019607843137}
\definecolor{color2}{rgb}{0.172549019607843,0.627450980392157,0.172549019607843}
\definecolor{color3}{rgb}{0.83921568627451,0.152941176470588,0.156862745098039}
\definecolor{color4}{rgb}{0.580392156862745,0.403921568627451,0.741176470588235}

\begin{axis}[
legend cell align={left},
legend style={
  fill opacity=0.8,
  draw opacity=1,
  text opacity=1,
  at={(0.97,0.03)},
  anchor=south east,
  draw=white!80!black
},
tick align=outside,
tick pos=left,
x grid style={white!69.0196078431373!black},
scaled x ticks={real:1000},
xtick scale label code/.code={},
xlabel={Gflops},
xmin=900, xmax=4800,
xtick style={color=black},
y grid style={white!69.0196078431373!black},
scaled y ticks={real:0.01},
ytick scale label code/.code={},
ylabel={accuracy (\%)},
ymin=0.72, ymax=0.82,
ytick style={color=black}
]

\addplot [color1, mark=triangle, mark options={}]
table {%
1082.99 0.723573
1373.18 0.749573
1706.49 0.775748
2286.83 0.797683
2519.18 0.801368
2805.78 0.8048
3274.05 0.80658
4673.08 0.812793
4673.29 0.811678
};
\addlegendentry{K-Means}
\addplot [color2, dashed, mark=x, mark options={solid}]
table {%
4672.4 0.8131
3273.31 0.808483
2519.07 0.801998
2287.66 0.79826
1706.29 0.775963
1373.52 0.750963
1082.88 0.722635
};
\addlegendentry{K-Means, rand cluster init}
\addplot [color3, mark=triangle, mark options={solid,rotate=180}]
table {%
996.607 0.723278
1258.82 0.753863
1571.34 0.778173
2131.48 0.798708
2345.74 0.801208
2610.62 0.805418
3037.95 0.807448
4438.79 0.81379
};
\addlegendentry{K-Medoids}
\addplot [color4, dashed, mark=x, mark options={solid,rotate=180}]
table {%
4438.79 0.811148
3037.95 0.806785
2610.62 0.806078
2345.74 0.80066
2131.48 0.797863
1571.34 0.77726
1258.82 0.75128
996.607 0.723445
};
\addlegendentry{K-Medoids, rand cluster init}
\end{axis}

\end{tikzpicture}
    \caption{Default initialization \vs random initialization. Our Token Pooling is robust to initialization of clustering algorithms. The default initialization is top-K \wrt significance score, see \autoref{al:downsampling}.}
    \label{fig:my_label}
\end{figure}
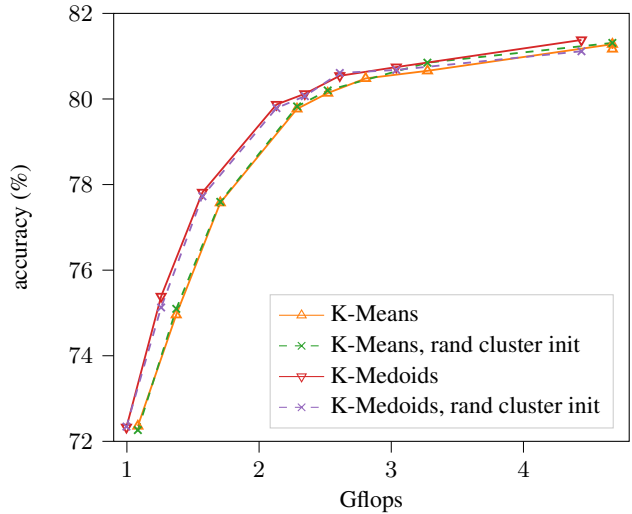

We examine the effect of the cluster center initialization. 
We compare our default initialization, which uses the tokens with top-K significance scores as initial cluster centers, with random initialization, which randomly selects tokens as initial cluster centers.
As shown in \autoref{fig:my_label}, Token Pooling is robust to the initialization methods.

\section{Training details}
\label{app:2x}

To fairly compare PoWER-BERT and the baseline methods with the proposed Token Pooling, all methods (except convolution downsampling) use the same target number of  tokens for downsampling layers after each transformer block.
Specifically, we run the second stage of the PoWER-BERT training for 30 epochs with various values of the token-selection weight parameter producing a family of models.
Each of the models has a different number of retained tokens at each of its $L$ transformer blocks: $\vK = (K_1, \dots, K_L)$.
\autoref{app:clustering ablations} lists all combinations of $\vK$.
We then finetune the base DeiT model (from the first stage training of PoWER-BERT) with the selected $\vK$ using the third (last) stage of PoWER-BERT, random selection, importance selection or our Token Pooling using the same $\vK$.

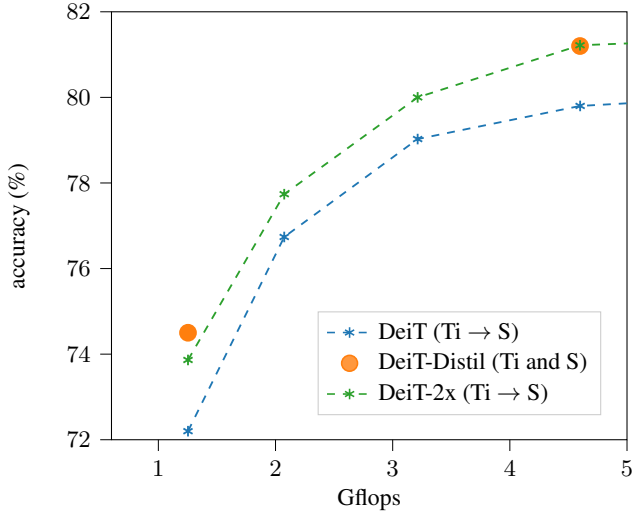
\begin{figure}[H]
    \centering
    \begin{tikzpicture}

\definecolor{color0}{rgb}{0.12156862745098,0.466666666666667,0.705882352941177}
\definecolor{color1}{rgb}{1,0.498039215686275,0.0549019607843137}
\definecolor{color2}{rgb}{0.172549019607843,0.627450980392157,0.172549019607843}

\begin{axis}[
legend cell align={left},
legend style={
  fill opacity=0.8,
  draw opacity=1,
  text opacity=1,
  at={(0.95,0.3)},
  draw=white!80!black
},
tick align=outside,
tick pos=left,
x grid style={white!69.0196078431373!black},
xlabel={Gflops},
xmin=0.600, xmax=5.000,
xtick style={color=black},
y grid style={white!69.0196078431373!black},
ylabel={accuracy (\%)},
ymin=72, ymax=82,
ytick style={color=black}
]
\addplot [semithick, color0, dashed, mark=asterisk, mark options={solid}]
table {%
1.253 72.2
2.0744 76.738
3.2131 79.026
4.600 79.8
17.600 81.8
};
\addlegendentry{DeiT (Ti $\rightarrow$ S)}
\addplot [semithick, color1, mark=*, mark size=3, mark options={solid}, only marks]
table {%
1.253 74.5
4.59888 81.2
17.600 83.4
};
\addlegendentry{DeiT-Distil (Ti and S)}
\addplot [semithick, color2, dashed, mark=asterisk, mark options={solid}]
table {%
1.253 73.864
2.07438 77.738
3.21306 80
4.59888 81.218
17.5638 82.584
};
\addlegendentry{DeiT-2x (Ti $\rightarrow$ S)}
\end{axis}

\end{tikzpicture}
    \caption{
    This figure shows the results of the pretrained DeiT models provided by \citet{touvron2020deit} (\mbox{\labelfont DeiT}) and the DeiT models trained with our protocol (\mbox{\labelfont DeiT-2x}). Our training protocal uses the same hyper-parameters provided by \citet{touvron2020deit}, but after the model is trained, we finetune the model using the same hyper-parameters (\ie restart the learning rate schedule). We also show \mbox{\labelfont DeiT-Distil} results (cited from \citep{touvron2020deit}), which use knowledge distillation.
    }
    \label{fig:deit 2x}
\end{figure}

We find that the DeiT models provided by \citet{touvron2020deit} are under-fit, and their accuracy improves with additional training, see \autoref{fig:deit 2x}. 
After the standard DeiT training, we restart the training. 
This ensures that downsampling models and DeiT with almost the same amount of training.

\section{Convolution downsampling}
\label{sec:convolution}

\begin{figure}[H]
    \centering
    \includegraphics{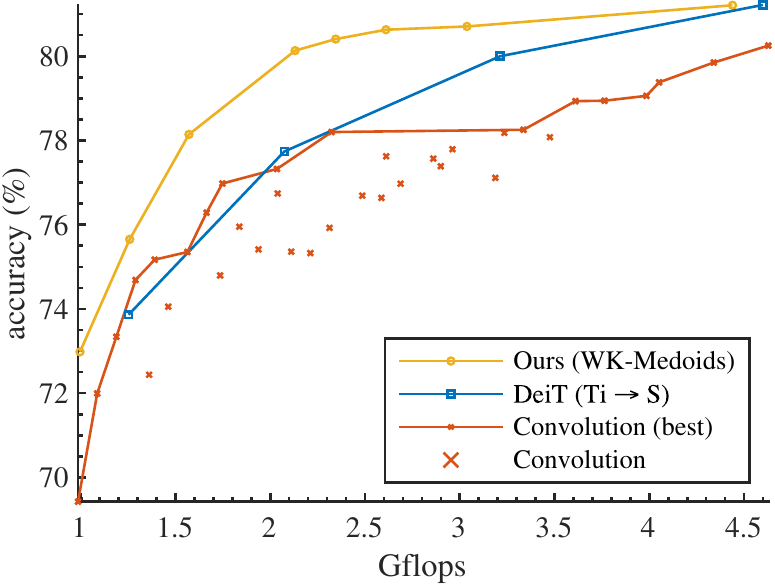}
    \caption{Results of convolution downsampling}
    \label{fig:conv all}
\end{figure}

As mentioned in the main paper, we enumerate combinations of layers to insert the convolution downsampling layer.
We use 2x2 convolution with stride 2 like in \citet{liu2021swin}.
To keep the feature dimensionality of DeiT the same (and evaluate the pure effect of the downsampling layers), the output feature dimensionality is the same as the input.
With 196 tokens in DeiT-S model, we can include no more than 3 convolution downsampling layers as each layer reduces the number of token by a factor of 4.
When using 3 layers at depths $l_1<l_2<l_3$, we restrict $l_2-l_1 = l_3 - l_2$.
With these constraints, we enumerate all possible downsampling configurations.
Each of the combinations produces a model with a different computation-accuracy trade-off, and we report the Pareto front, \ie, the best accuracy these models achieve at a given flop.
See \autoref{fig:conv all}.

\section{Results without finetuning}
\label{app:no tuning}

\begin{figure}[H]
    \centering
    \begin{tikzpicture}

\definecolor{color0}{rgb}{0.12156862745098,0.466666666666667,0.705882352941177}
\definecolor{color1}{rgb}{1,0.498039215686275,0.0549019607843137}
\definecolor{color2}{rgb}{0.172549019607843,0.627450980392157,0.172549019607843}
\definecolor{color3}{rgb}{0.83921568627451,0.152941176470588,0.156862745098039}
\definecolor{color4}{rgb}{0.580392156862745,0.403921568627451,0.741176470588235}
\definecolor{color5}{rgb}{0.549019607843137,0.337254901960784,0.294117647058824}
\definecolor{color6}{rgb}{0.890196078431372,0.466666666666667,0.76078431372549}
\definecolor{color7}{rgb}{0.92, 0.69, 0.12}

\begin{axis}[
legend cell align={left},
legend style={
  fill opacity=0.8,
  draw opacity=1,
  text opacity=1,
  at={(0.97,0.03)},
  anchor=south east,
  draw=white!80!black
},
tick align=outside,
tick pos=left,
x grid style={white!69.0196078431373!black},
scaled x ticks={real:1000},
xtick scale label code/.code={},
xlabel={\(\displaystyle \cdot 10^9\), Flops},
xmin=2000, xmax=4800,
xtick style={color=black},
y grid style={white!69.0196078431373!black},
scaled y ticks={real:0.01},
ytick scale label code/.code={},
ylabel={accuracy (\%)},
ymin=0.2, ymax=0.82,
ytick style={color=black}
]
\addplot [semithick, color0, mark=asterisk, mark options={solid}, only marks]
table {%
4598.88 0.81218
};
\addlegendentry{DeiT-S}
\addplot [semithick, color1, mark=square, mark options={solid}]
table {%
1070.4 0.00102
1348.92 0.00132
1677.09 0.00162
2250.52 0.30088
2475.84 0.53544
2752.11 0.64056
3199.84 0.74836
4614.79 0.79826
};
\addlegendentry{K-Means, no finetuning}
\addplot [semithick, color2, mark=o, mark options={solid}]
table {%
1062.77 0.00106
1337.04 0.0014
1662.65 0.00172
2242.95 0.41814
2466.77 0.59782
2741.19 0.6711
3188.96 0.75276
4608.22 0.79824
};
\addlegendentry{WK-Means, no finetuning}
\addplot [semithick, color3, mark=diamond, mark options={solid}]
table {%
1065.58 0.00124
1338.63 0.002
1664.93 0.00432
2247.78 0.67906
2472.68 0.7163
2749.53 0.74604
3200.72 0.77164
4618.07 0.79784
};
\addlegendentry{WK-Means, carry, no finetuning}
\addplot [semithick, color4, mark=x, mark options={solid}]
table {%
996.607 0.0011
1258.82 0.00122
1571.34 0.00166
2131.48 0.28944
2345.74 0.52724
2610.62 0.63766
3037.95 0.74798
4438.79 0.79828
};
\addlegendentry{K-Medoids, no finetuning}
\addplot [semithick, color6, mark=triangle, mark options={solid}]
table {%
996.607 0.00102
1258.82 0.00136
1571.34 0.00178
2131.48 0.41886
2345.74 0.59428
2610.62 0.67114
3037.95 0.75228
4438.79 0.79816
};
\addlegendentry{WK-Medoids, no finetuning}
\addplot [semithick, color7, mark=+, mark options={solid}]
table {%
996.607 0.00154
1258.82 0.00168
1571.34 0.00448
2131.48 0.68004
2345.74 0.71846
2610.62 0.74474
3037.95 0.77202
4438.79 0.79788
};
\addlegendentry{WK-Medoids, carry, no finetuning}
\end{axis}

\end{tikzpicture}
    \caption{The figure shows the performance results when we insert Token Pooling layers into a pretrained DeiT-S \textit{without} finetuning the model (\ie skip step 3 of the training protocol in \autoref{sec:results}). 
    }
    \label{fig:no finetuning}
\end{figure}
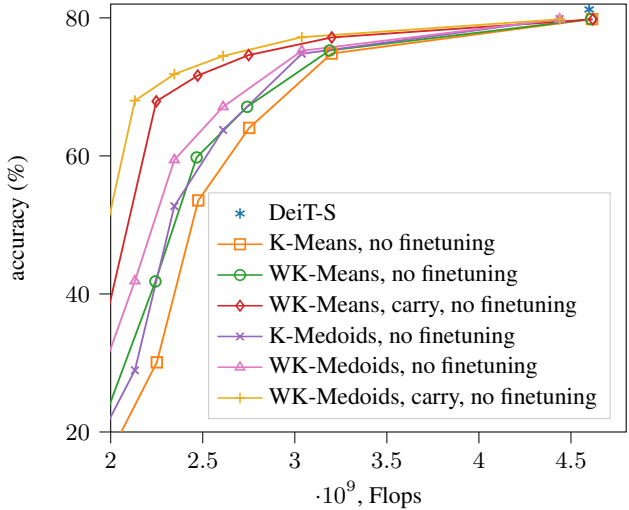

Since Token Pooling minimizes the reconstruction error due to token downsampling, in this section, we evaluate the performance when we insert Token Pooling layers into a pre-trained model \textit{without} finetuning. 
\autoref{fig:no finetuning} shows the results when we directly insert Token Pooling layers (using the same downsampling ratios in \autoref{fig:token poolings}) into a pretrained DeiT-S. 
As can be seen, minimizing the reconstruction error, Token Pooling preserves information that enables the model to retain accuracy during token downsampling.

In \autoref{fig:no finetuning}, we also show the results when we replace tokens to their cluster centers (WK-Means, carry, no finetuning and WK-Medoids, carry, no finetuning). 
Specifically, in addition to outputting $K$ cluster centers, we count the number of tokens assigned to a cluster center and \textit{carry} the count when we compute softmax-attention in the next transformer blocks.
This operation preserves the attention weights, and since the models are not trained after inserting Token Pooling layers, it preserves the most accuracy. 
In practice, when the models are trained with Token Pooling layers, the carry operation is not important.

\section{Normalized key and query vectors}
\label{app:normalized key query}

\begin{figure}
	\centering
		\begin{subfigure}[t]{0.48\linewidth}
		\centering
		\includegraphics{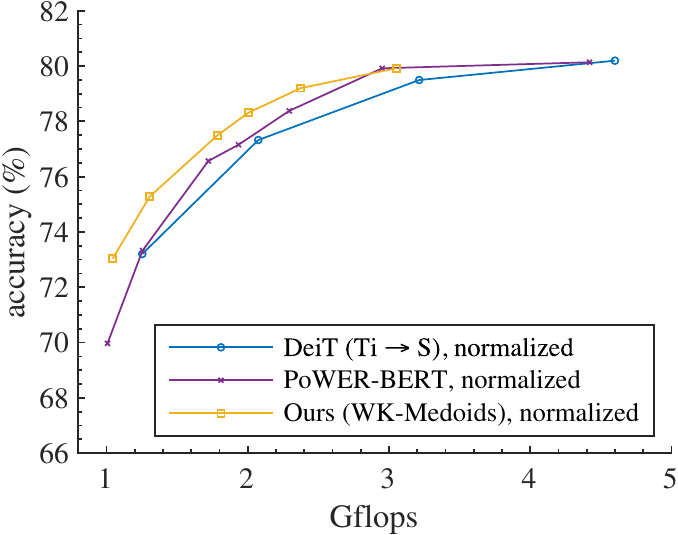}
		\caption{$\alpha = 1$}
		\label{fig:normalized results norm 1}
	\end{subfigure}
	\hfill
	\begin{subfigure}[t]{0.48\linewidth}
		\centering
		\includegraphics{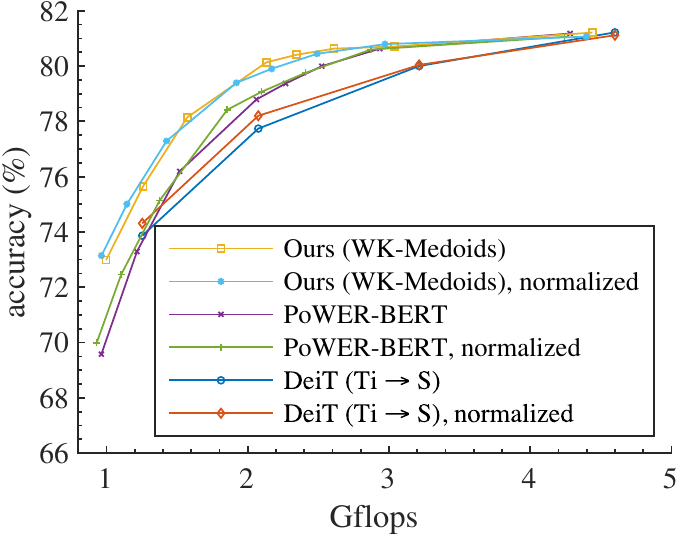}
		\caption{Learned $\alpha$ for each layer}
		\label{fig:normalized results learned norm}
	\end{subfigure}
	\vspace{-3mm}
	\caption{Results of models using normalized key and query vectors with (a) $\alpha = 1$ and (b) learned $\alpha$ in \eqref{eq:attention}. The base model architecture is DeiT-S.}
	\label{fig:normalized results}
\end{figure}

Our analysis of softmax-attention in \autoref{sec:low-pass} assumes the query and the key vectors are normalized to have a constant $\ell^2$ norm.
It has been observed by \citet{kitaev2020reformer} that normalizing key and query vectors does not change the performance of a transformer. 
Thus, for all experiments, we train models \textit{without} the norm normalization. 
In this section, we verify this observation by training various DeiT, PoWER-BERT, and Token Pooling \textit{with} normalized keys and queries.
We found that the scalar $\alpha$ in the softmax attention layer \eqref{eq:attention} can affect the performance of a transformer with normalized keys and queries.
As can be seen in \autoref{fig:normalized results norm 1}, setting the scalar $\alpha = 1$ slightly deteriorates the performance of the model. 
Instead of using a fixed $\alpha$, we let the model learn the $\alpha$ for each layer.
As can be seen in \autoref{fig:normalized results learned norm}, learning $\alpha$ enables the resulting models to achieve similar cost-accuracy trade-off as the standard (unnormalized) models.
With or without the normalization and the learnable $\alpha$, the proposed Token Pooling significantly improves the cost-accuracy trade-off of DeiT  and outperforms PoWER-BERT.

\section{Clustering algorithm ablations}
\label{app:clustering ablations}

\begin{figure}[t]
	\centering
	\vspace{3mm}
	\begin{subfigure}[t]{0.48\linewidth}
		\centering
		\includegraphics[width=\linewidth]{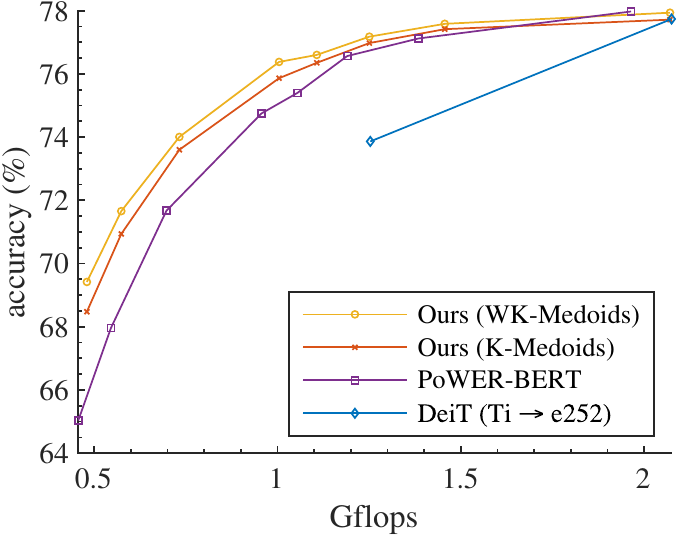}
		\caption{DeiT-e252 as the base architecture}
	\end{subfigure}
	\hfill
	\begin{subfigure}[t]{0.48\linewidth}
		\centering
		\includegraphics[width=\linewidth]{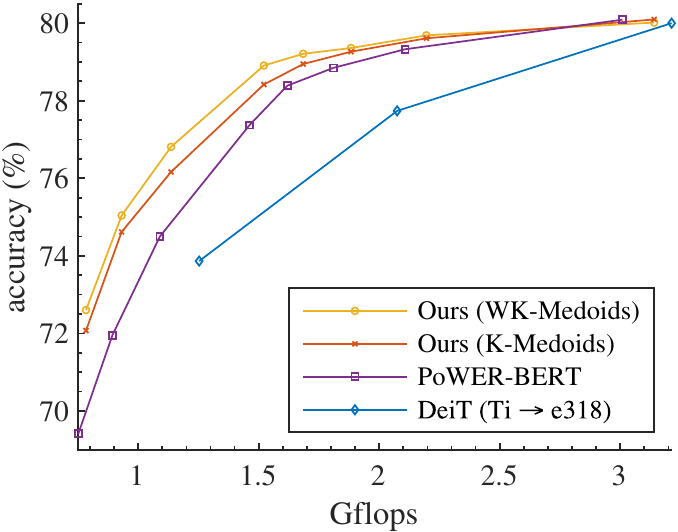}
		\caption{DeiT-e318 as the base architecture}
	\end{subfigure}
	\caption{This figure compares the cost-accuracy curves of Token Pooling with PoWER-BERT using (a) DeiT-e252  and (b) DeiT-e318 as the base architectures.}
	\label{fig:e252 e318}
\end{figure}

Tables \ref{tab:clustering ablations}--\ref{tab:data best} detail the results of PoWER-BERT and Token Pooling on the DeiT architectures that we tested (DeiT-S, DeiT-e318, and DeiT-e252).
\autoref{fig:e252 e318} shows the results.
Table \ref{tab:data best} details the results of the best cost-accuracy trade-off achieved by the proposed Token Pooling (using K-Medoids and WK-Medoids) and PoWER-BERT via varying token sparsity and feature dimensionality of DeiT.

Apart from the standard K-Means and K-Medoids, other clustering approaches could be used. 
Many methods are not suitable due to efficiency constraints. 
For example, normalized cut \citep{shi2000normalized} uses expensive spectral methods. 
One prerequisite of using K-Means and K-Medoids is the number of clusters, $K$. 
While selecting $K$ for each of the layers may be tedious and difficult, one can choose $K$ via computational budgets and heuristics. 
In this work, we use the automatic search procedure proposed by \citet{goyal2020power} to determine $K$, see \autoref{app:2x}. 

Since both K-Means and K-Medoids require specifying the number of clusters $K$ in advance, one may consider using methods automatically determining $K$. 
Nevertheless, such methods typically have other parameters, which are less interpretable than $K$. 
For example, mean-shift \citep{cheng1995mean} or quick-shift \citep{vedaldi2008quick} require specifying the kernel size. 
From our experience, determining these parameters is challenging.
Also, since the number of clusters (and hence the number of output tokens) is determined on the fly during inference, the computational requirement can fluctuate, making deployment of these models difficult.

\begin{table}
	\centering
	\def\colwidth{16mm} 
	\setlength{\tabcolsep}{5pt}
	\renewcommand{\arraystretch}{1.2}
	\begin{adjustbox}{max width=\textwidth}
	\begin{tabular}{cccccccc}
		\toprule
		\multicolumn{2}{c}{DeiT (Ti $\rightarrow$ S)} &
		\multicolumn{2}{c}{PoWER-BERT} & 
		\multicolumn{2}{c}{Token Pooling (K-Medoids)} &
		\multicolumn{2}{c}{Token Pooling (WK-Medoids)}  \\
		\makecell[c]{Gflops} & \makecell[c]{Accuracy (\%)} & 
		\makecell[c]{Gflops} & \makecell[c]{Accuracy (\%)} & 
		\makecell[c]{Gflops} & \makecell[c]{Accuracy (\%)} & 
		\makecell[c]{Gflops} & \makecell[c]{Accuracy (\%)} \\
		\cmidrule(lr){1-2}
		\cmidrule(lr){3-4}
		\cmidrule(lr){5-6}
		\cmidrule(lr){7-8}
		-& -& %
		0.46 & 65.0 & %
		0.48 & 68.5 & %
		0.48 & \textbf{69.4}  %
		\\ 
		-& -& %
		0.55 & 68.0 & %
		0.57 & 70.9 & %
		0.57 & \textbf{71.7}  %
		\\ 
		-& -& %
		0.70 & 71.7 & %
		0.73 & 73.6 & %
		0.73 & \textbf{74.0 } %
		\\ 
		-& -& %
		0.89 & 72.0 & %
		0.93 & 74.6 & %
		0.93 & \textbf{75.0}  %
		\\ 
		-& -& %
		0.96 & 74.8 & %
		1.00 & 75.9 & %
		1.00 & \textbf{76.4 } %
		\\ 
		-& -& %
		1.05 & 75.4 & %
		1.11 & 76.4 & %
		1.11 & \textbf{76.6}  %
		\\ 
		1.25 & 73.9 & %
		1.19 & 76.6 & %
		1.25 & 77.0 & %
		1.25 & \textbf{77.2 } %
		\\ 
		-& -& %
		1.38 & 77.1 & %
		1.46 & 77.4 & %
		1.46 & \textbf{77.6}  %
		\\ 
		-& -& %
		1.46 & 77.4 & %
		1.52 & 78.4 & %
		1.52 & \textbf{78.9}  %
		\\ 
		-& -& %
		1.62 & 78.4 & %
		1.68 & 78.9 & %
		1.68 & \textbf{79.2}  %
		\\ 
		2.07 & 77.7 & %
		1.81 & 78.8 & %
		1.88 & 79.3 & %
		1.88 & \textbf{79.4 } %
		\\ 
		-& -& %
		2.11 & 79.3 & %
		2.13 & 79.9 & %
		2.13 &\textbf{ 80.1 } %
		\\ 
		-& -& %
		2.27 & 79.4 & %
		2.35 & 80.1 & %
		2.35 & \textbf{80.4 } %
		\\ 
		-& -& %
		2.52 & 80.0 & %
		2.61 & 80.5 & %
		2.61 & \textbf{80.6}  %
		\\ 
		3.21 & 80.0 & %
		2.93 & 80.6 & %
		3.04 & \textbf{80.7} & %
		3.04 & \textbf{80.7}  %
		\\ 
		4.60 & 81.2 & %
		4.28 & 81.2 & %
		4.44 & \textbf{81.4 }& %
		4.44 & 81.2  %
		\\ 
		\bottomrule
	\end{tabular}
	\end{adjustbox}
	\caption{Best cost-accuracy trade-off achieved  by PoWER-BERT and the proposed Token Pooling via varying sparsity level and feature dimensionality. }
	\label{tab:data best}
\end{table}

\begin{table}[t]
	\centering
	\renewcommand{\arraystretch}{1.05}
		\begin{adjustbox}{max width=\textwidth}
		\begin{tabular}{c|c||c|c}
			clustering method & Weighting & ImageNet Accuracy & GFlops \\\hline\hline
			\multicolumn{4}{l}{Sparsity level 0: $\bm K=[196, 196, 195, 194, 189, 180, 173, 173, 173, 173, 173, 173]$} \\ \hline
			PoWER-BERT & N/A & 81.2 & 4.3 \\ \hline
			
			\multirow{2}{*}{K-Means} &  & 81.3 & 4.7 (+0.4) \\ 
			& \checkmark & 81.1 &  4.7  (+0.4) \\\hline
			
			\multirow{2}{*}{K-Medoids} & &  \textbf{81.4} &  4.4 (+0.1) \\ 
			& \checkmark & 81.2 & 4.4 (+0.1) \\ \hline\hline

			\multicolumn{4}{l}{Sparsity level 1: $\bm K=[196, 195, 193, 188, 169, 140, 121, 110, 73, 38, 7, 0]$} \\ \hline
			PoWER-BERT & N/A & 80.6 & 2.9  \\ \hline
			
			\multirow{2}{*}{K-Means} &  & 80.7 & 3.3 (+0.4) \\ 
			& \checkmark & \textbf{80.8} & 3.3  (+0.4) \\\hline
			
			\multirow{2}{*}{K-Medoids} & & 80.7 & 3.0  (+0.1)\\ 
			& \checkmark & 80.7 &  3.0 (+0.1) \\ \hline\hline

			\multicolumn{4}{l}{Sparsity level 2: $\bm K=[196, 195, 190, 177, 141, 108, 84, 69, 35, 18, 3, 0]$} \\ \hline
			PoWER-BERT & N/A & 80.0 & 2.5 \\ \hline
			
			\multirow{2}{*}{K-Means} &  & 80.5 & 2.8 (+0.3) \\ 
			& \checkmark & 80.5 & 2.8 (+0.3)\\\hline
			
			\multirow{2}{*}{K-Medoids} & & 80.5 & 2.6 (+0.1) \\ 
			& \checkmark & \textbf{80.6} & 2.6 (+0.1)  \\ \hline\hline

			\multicolumn{4}{l}{Sparsity level 3: $\bm K=[196, 194, 187, 163, 118, 85, 58, 47, 20, 12, 2, 0]$} \\ \hline
			PoWER-BERT & N/A & 79.4 & 2.3  \\ \hline
			
			\multirow{2}{*}{K-Means} &  & 80.1 & 2.5 (+0.2) \\ 
			& \checkmark & \textbf{80.4 }& 2.5 (+0.2) \\\hline
			
			\multirow{2}{*}{K-Medoids} && 80.1 & 2.3 (+0.08)\\ 
			& \checkmark & \textbf{80.4} & 2.3 (+0.08) \\ \hline\hline

			\multicolumn{4}{l}{Sparsity level 4: $\bm K=[196, 193, 179, 142, 97, 64, 46, 34, 13, 9, 1, 0]$} \\ \hline
			PoWER-BERT & N/A & 78.8 & 2.1 \\ \hline
			
			\multirow{2}{*}{K-Means} &  & 79.8 & 2.3 (+0.2) \\ 
			& \checkmark & 80.0 & 2.3 (+0.2) \\\hline
			
			\multirow{2}{*}{K-Medoids} & & 79.9 & 2.1 (+0.07) \\ 
			& \checkmark & \textbf{80.1} & 2.1 (+0.07) \\ \hline\hline

			\multicolumn{4}{l}{Sparsity level 5: $\bm K=[194, 183, 142, 89, 41, 20, 10, 7, 0, 0, 0, 0]$} \\ \hline
			PoWER-BERT & N/A & 76.2 & 1.5 \\ \hline
			
			\multirow{2}{*}{K-Means} &  & 77.6 & 1.7 (+0.2) \\ 
			& \checkmark & \textbf{78.1} & 1.7 (+0.2) \\\hline
			
			\multirow{2}{*}{K-Medoids} &  & 77.8 & 1.6 (+0.05) \\ 
			& \checkmark & \textbf{78.1} & 1.6 (+0.05) \\ \hline\hline

			\multicolumn{4}{l}{Sparsity level 6: $\bm K=[186, 162, 102, 56, 13, 4, 2, 2, 0, 0, 0, 0]$} \\ \hline
			PoWER-BERT & N/A & 73.3 & 1.2 \\ \hline
			
			\multirow{2}{*}{K-Means} &  & 75.0 & 1.4 (+0.2) \\ 
			& \checkmark & 75.6 & 1.4 (+0.2) \\\hline
			
			\multirow{2}{*}{K-Medoids} & & 75.4 & 1.3 (+0.04) \\ 
			& \checkmark & \textbf{75.7} & 1.3 (+0.04) \\ \hline\hline

			\multicolumn{4}{l}{Sparsity level 7: $\bm K=[162, 129, 66, 33, 4, 1, 1, 0, 0, 0, 0, 0]$} \\ \hline
			PoWER-BERT & N/A & 69.6 & 1.0 \\ \hline
			
			\multirow{2}{*}{K-Means} &  & 72.4 & 1.1 (+0.1) \\ 
			& \checkmark & \textbf{73.0} & 1.1 (+0.1) \\\hline
			
			\multirow{2}{*}{K-Medoids} & & 72.3 & 1.0 (+0.03) \\ 
			& \checkmark & \textbf{73.0} & 1.0 (+0.03) \\ \hline\hline
		\end{tabular}
	\end{adjustbox}
	\caption{Results of applying Token Pooling and PoWER-BERT on \textbf{DeiT-S} model. The models are grouped by $\mK$ described in \autoref{app:2x}. The integer list denotes the maximal number of tokens retained after each transformer block. These numbers do not take into account the classification token, which is always retained. Thus, ``0'' means that only the classification token remains. Additional flops (denoted by the parentheses) are due to clustering.}
	\label{tab:clustering ablations}
\end{table}

\begin{table}[t]
	\centering
	\renewcommand{\arraystretch}{1.2}
	\begin{adjustbox}{max width=\textwidth}
		\begin{tabular}{c|c||c|c}
			clustering method & Weighting & ImageNet Accuracy & GFlops \\\hline\hline
			\multicolumn{4}{l}{Sparsity level 0: $\bm K=[196, 196, 196, 194, 192, 184, 181, 181, 170, 170, 170, 170]$} \\ \hline
			PoWER-BERT & N/A & 80.1 & 3.0 \\ \hline
				
			\multirow{2}{*}{K-Medoids} & 
			&  \textbf{80.1} & 3.1 (+0.1)\\ 
			& \checkmark & 80.0 & 3.1 (+0.1) \\ \hline\hline

			\multicolumn{4}{l}{Sparsity level 1: $\bm K=[196, 195, 195, 190, 171, 147, 132, 122, 66, 43, 15, 0]$} \\ \hline
			PoWER-BERT & N/A & 79.3 & 2.1  \\ \hline
			
			\multirow{2}{*}{K-Medoids} & 
			& 79.6 & 2.2 (+0.1) \\ 
			& \checkmark & \textbf{79.7} &  2.2 (+0.1) \\ \hline\hline

			\multicolumn{4}{l}{Sparsity level 2: $\bm K=[196, 195, 193, 180, 143, 109, 92, 77, 40, 21, 4, 0]$} \\ \hline
			PoWER-BERT & N/A & 78.8 & 1.8 \\ \hline
				
			\multirow{2}{*}{K-Medoids} & 
			& 79.3 & 1.9 (+0.09) \\ 
			& \checkmark & \textbf{79.4} & 1.9 (+0.09) \\ \hline\hline

			\multicolumn{4}{l}{Sparsity level 3: $\bm K=[196, 195, 190, 165, 119, 84, 65, 50, 26, 14, 3, 0]$} \\ \hline
			PoWER-BERT & N/A & 78.3 & 1.6  \\ \hline
			
			\multirow{2}{*}{K-Medoids} &
			& 78.9 & 1.7 (+0.07) \\ 
			& \checkmark & \textbf{79.2} & 1.7 (+0.07) \\ \hline\hline

			\multicolumn{4}{l}{Sparsity level 4: $\bm K=[196, 194, 184, 149, 97, 65, 47, 33, 12, 9, 2, 0]$} \\ \hline
			PoWER-BERT & N/A & 77.4 & 1.5 \\ \hline
				
			\multirow{2}{*}{K-Medoids} & 
			& 78.4 & 1.5 (+0.06) \\ 
			& \checkmark & \textbf{78.9 } & 1.5 (+0.06) \\ \hline\hline

			\multicolumn{4}{l}{Sparsity level 5: $\bm K=[196, 186, 153, 93, 40, 15, 12, 9, 0, 0, 0, 0]$} \\ \hline
			PoWER-BERT & N/A & 74.5 & 1.1 \\ \hline
				
			\multirow{2}{*}{K-Medoids} &  
			& 76.2 & 1.1 (+0.05) \\ 
			& \checkmark & \textbf{76.8} & 1.1 (+0.05) \\ \hline\hline

			\multicolumn{4}{l}{Sparsity level 6: $\bm K=[193, 173, 109, 52, 16, 4, 4, 4, 0, 0, 0, 0]$} \\ \hline
			PoWER-BERT & N/A & 72.0 & 0.89 \\ \hline
			
			\multirow{2}{*}{K-Medoids} & 
			& 74.6 & 0.93 (+0.04) \\ 
			& \checkmark & \textbf{75.0} & 0.93 (+0.04) \\ \hline\hline

			\multicolumn{4}{l}{Sparsity level 7: $\bm K=[183, 145, 80, 33, 5, 1, 1, 2, 0, 0, 0, 0]$} \\ \hline
			PoWER-BERT & N/A & 69.4 & 0.75 \\ \hline
			
			\multirow{2}{*}{K-Medoids} & 
			& 72.1 & 0.78 (+0.03) \\ 
			& \checkmark & \textbf{72.6} & 0.78 (+0.03) \\ \hline\hline
		\end{tabular}
	\end{adjustbox}
	\caption{Results of applying Token Pooling and PoWER-BERT on \textbf{DeiT-e318} model. The models are grouped by $\mK$ described in \autoref{app:2x}. The integer list denotes the maximal number of tokens retained after each transformer block. These numbers do not take into account the classification token, which is always retained. Thus, ``0'' means that only the classification token remains. Additional flops (denoted by the parentheses) are due to clustering.}
	\label{tab:clustering ablations deit-e318}
\end{table}

\begin{table}[t]
	\centering
	\renewcommand{\arraystretch}{1.2}
	\begin{adjustbox}{max width=\textwidth}
		\begin{tabular}{c|c||c|c}
			clustering method & Weighting & ImageNet Accuracy & GFlops \\\hline\hline
			\multicolumn{4}{l}{Sparsity level 0: $\bm K=[196, 196, 196, 195, 193, 189, 177, 177, 177, 177, 177, 177]$} \\ \hline
			PoWER-BERT & N/A & \textbf{78.0} & 2.0 \\ \hline
			
			\multirow{2}{*}{K-Medoids} & 
			&  77.7 &  2.1 (+0.1)\\ 
			& \checkmark & 77.9 & 2.1 (+0.1) \\ \hline\hline

			\multicolumn{4}{l}{Sparsity level 1: $\bm K=[196, 196, 195, 189, 176, 161, 123, 122, 76, 48, 17, 0]$} \\ \hline
			PoWER-BERT & N/A & 77.1 & 1.4  \\ \hline
			
			\multirow{2}{*}{K-Medoids} & 
			& 77.4 & 1.5 (+0.07) \\ 
			& \checkmark & \textbf{77.6} &  1.5 (+0.07) \\ \hline\hline

			\multicolumn{4}{l}{Sparsity level 2: $\bm K=[196, 196, 193, 179, 154, 123, 88, 79, 39, 22, 5, 0],$} \\ \hline
			PoWER-BERT & N/A & 76.6 & 1.2 \\ \hline
			
			\multirow{2}{*}{K-Medoids} & 
			& 77.0 & 1.3 (+0.06) \\ 
			& \checkmark & \textbf{77.2} & 1.3 (+0.06) \\ \hline\hline

			\multicolumn{4}{l}{Sparsity level 3: $\bm K=[196, 195, 188, 167, 131, 96, 63, 52, 13, 11, 2, 0],$} \\ \hline
			PoWER-BERT & N/A & 75.4 & 1.1  \\ \hline
			
			\multirow{2}{*}{K-Medoids} &
			& 76.4 & 1.1 (+0.05) \\ 
			& \checkmark & \textbf{76.6} & 1.1 (+0.05) \\ \hline\hline

			\multicolumn{4}{l}{Sparsity level 4: $\bm K=[196, 194, 181, 147, 111, 75, 48, 36, 5, 7, 2, 0]$} \\ \hline
			PoWER-BERT & N/A & 74.8 & 0.96 \\ \hline
			
			\multirow{2}{*}{K-Medoids} & 
			& 75.9 & 1.0 (+0.04) \\ 
			& \checkmark & \textbf{76.4 } & 1.0 (+0.04) \\ \hline\hline

			\multicolumn{4}{l}{Sparsity level 5: $\bm K=[196, 187, 129, 88, 54, 20, 10, 10, 0, 1, 0, 0]$} \\ \hline
			PoWER-BERT & N/A & 71.7 & 0.70 \\ \hline
			
			\multirow{2}{*}{K-Medoids} &  
			& 73.6 & 0.73 (+0.03) \\ 
			& \checkmark & \textbf{74.0} & 0.73 (+0.03) \\ \hline\hline

			\multicolumn{4}{l}{Sparsity level 6: $\bm K=[194, 163, 83, 44, 22, 6, 1, 3, 0, 0, 0, 0]$} \\ \hline
			PoWER-BERT & N/A & 68.0 & 0.55 \\ \hline
			
			\multirow{2}{*}{K-Medoids} & 
			& 70.9 & 0.57 (+0.02) \\ 
			& \checkmark & \textbf{71.7} & 0.57 (+0.02) \\ \hline\hline

			\multicolumn{4}{l}{Sparsity level 7: $\bm K=[188, 122, 58, 28, 12, 2, 0, 2, 0, 0, 0, 0]$} \\ \hline
			PoWER-BERT & N/A & 65.0 & 0.46 \\ \hline
			
			\multirow{2}{*}{K-Medoids} & 
			& 68.5 & 0.48 (+0.02) \\ 
			& \checkmark & \textbf{69.4} & 0.48 (+0.02) \\ \hline\hline
		\end{tabular}
	\end{adjustbox}
	\caption{Results of applying Token Pooling and PoWER-BERT on \textbf{DeiT-e252} model. The models are grouped by $\mK$ described in \autoref{app:2x}. The integer list denotes the maximal number of tokens retained after each transformer block. These numbers do not take into account the classification token, which is always retained. Thus, ``0'' means that only the classification token remains. Additional flops (denoted by the parentheses) are due to clustering.}
	\label{tab:clustering ablations deit-e252}
\end{table}


\end{document}